\documentclass[11pt,a4paper]{article}
\usepackage[hyperref]{naaclhlt2019}
\usepackage{times}
\usepackage{latexsym}
\usepackage{amsmath}
\usepackage{booktabs}
\usepackage{graphicx}
\usepackage{multirow}
\usepackage{lipsum}
\usepackage{adjustbox}
\usepackage{caption}
\usepackage{qtree}
\usepackage{xpatch}
\xapptocmd\normalsize{%
 \abovedisplayskip=8pt plus 3pt minus 9pt
 \abovedisplayshortskip=0pt plus 3pt
 \belowdisplayskip=8pt plus 3pt minus 9pt
 \belowdisplayshortskip=7pt plus 3pt minus 4pt
}{}{}

\usepackage{url}

\DeclareMathOperator*{\argmax}{arg\,max}
\DeclareMathOperator*{\softmax}{softmax}

\aclfinalcopy % Uncomment this line for the final submission
\def\aclpaperid{1399} %  Enter the acl Paper ID here

\title{Factorising AMR generation through syntax}

\author{Kris Cao \and Stephen Clark \\
  Computer Laboratory \\
  University of Cambridge \\
  United Kingdom \\
  {\tt \{kc391,sc609\}@cam.ac.uk}}

\date{}

\begin{document}
\maketitle
\begin{abstract}
  Generating from Abstract Meaning Representation (AMR) is an underspecified problem, as many syntactic decisions are not constrained by the semantic graph. To explicitly account for this underspecification, we break down generating from AMR into two steps: first generate a syntactic structure, and then generate the surface form. We show that decomposing the generation process this way leads to state-of-the-art single model performance generating from AMR without additional unlabelled data. We also demonstrate that we can generate meaning-preserving syntactic paraphrases of the same AMR graph, as judged by humans.
\end{abstract}

\section{Introduction}

Abstract Meaning Representation (AMR) \citep{Baranescu:13} is a semantic annotation framework which abstracts away from the surface form of text to capture the core `who did what to whom' structure. As a result, generating from AMR is underspecified (see Figure \ref{fig:amr} for an example). Single-step approaches to AMR generation \citep{Flanigan:16,Konstas:17,Song:16,Song:17} therefore have to decide the syntax and surface form of the AMR realisation in one go. We instead explicitly try and capture this syntactic variation and factor the generation process through a syntactic representation \citep{Walker:03,Dusek:16,Gardent:17,Currey:18}. 

First, we generate a delexicalised constituency structure from the AMR graph using a syntax model. Then, we fill out the constituency structure with the semantic content in the AMR graph using a lexicalisation model to generate the final surface form. Breaking down the AMR generation process this way provides us with several advantages: we disentangle the variance caused by the choice of syntax from that caused by the choice of words. We can therefore realise the same AMR graph with a variety of syntactic structures by sampling from the syntax model, and deterministically decoding using the lexicalisation model. We hypothesise that this generates better paraphrases of the reference realisation than sampling from a single-step model.

We linearise both the AMR graphs \citep{Konstas:17} and constituency trees \citep{Vinyals:15} to allow us to use sequence-to-sequence models \citep{Sutskever:14,Bahdanau:15} for the syntax and lexicalisation models. Further, as the AMR dataset is relatively small, we have issues with data sparsity causing poor parameter estimation for rarely seen words. We deal with this by anonymizing named entities, and including a \textit{copy mechanism} \citep{Vinyals:15b,See:17,Song:18} into our decoder, which allows open-vocabulary token generation. 

% We would like to use sequence-to-sequence models \citep{Sutskever:14,Bahdanau:15} in both steps of our generation procedure. However, AMR is naturally graph-structured. While we could modify the sequence-to-sequence framework to use graph-structured encoders \citep{Kipf:16,Song:18,Beck:18,Marcheggiani:18}, for ease of implementation we instead linearise the AMR graph as in \citet{Konstas:17}.

We show that factorising the generation process in this way leads to improvements in AMR generation, setting a new state of the art for single-model AMR generation performance training only on labelled data. We also verify our diverse generation hypothesis with a human annotation study.

%There has been recent work on controlling attributes of model-generated text, such as tense and sentiment \citep{Hu:17,Ficler:17}. We note that we have very fine control over aspects of the generated text, as between the AMR graph and the syntax tree we can essentially control every aspect of the generated sentence. 

% Our approach towards learning syntactic frames for predicate-argument tuples bears some resemblance to verb subcategorisation frame induction \citep{Manning:93,Preiss:07,Lippincott:12}. That task consists of building a lexicon of subcategorisation frames for each verb in a corpus of text. A subcategorisation frame represents the syntactic structure of arguments that a verb can take: for example, the two sentences in the introductory paragraph demonstrate that \textigt{give} has two frames, one with two direct objects and one with a direct and an indirect object. However, our task differs in that we are already given annotated predicate-argument structure, rather than having to induce this from raw text.

\section{Data}
\label{sec:data}
\begin{table}
\centering
\begin{tabular}{|l|}
\toprule 
\texttt{(g / give-01} \\
\hspace{1em} \texttt{:ARG0 (i / I)} \\
\hspace{1em} \texttt{:ARG1 (b / ball)} \\
\hspace{1em} \texttt{:ARG2 (d / dog))} \\
\midrule
\texttt{\small give :arg0 i :arg1 ball :arg2 dog} \\
\midrule
I [gave]\textsubscript{VP} [the dog]\textsubscript{NP} [a ball]\textsubscript{NP} \\
I [gave]\textsubscript{VP} [the ball]\textsubscript{NP} [to a dog]\textsubscript{PP} \\
\bottomrule
\end{tabular}
\captionof{figure}{An example AMR graph, with variable names and verb senses, followed by the input to our system after preprocessing, and finally two sample realisations different in syntax.}
\label{fig:amr}
\vspace{-1em}

\end{table}

\paragraph{Abstract Meaning Repreentation} Abstract Meaning Representation is a semantic annotation formalism which represents the meaning of an English utterance as a rooted directed acyclic graph. Nodes in the graph represent entities, events, properties and states mentioned in the text, while leaves of the graph label the nodes with concepts (which do not have to be aligned to spans in the text). Re-entrant nodes correspond to coreferent entities. Edges in the graph represent relations between entities in the text. See Figure \ref{fig:amr} for an example of an AMR graph, together with sample realisations.

\citet{Konstas:17} outline a set of preprocessing procedures for AMR graphs to both render them suitable for sequence-to-sequence learning and to ameliorate data sparsity; we follow the same pipeline. We train our models on the two most recent AMR releases. LDC2017T10 has roughly 36k training sentences, while LDC2015E86 is about half this size. Both share dev and test sets, facilitating comparison.

\paragraph{Constituency syntax} While there are many syntactic annotation formalisms, we use delexicalised Penn treebank-style constituency trees to represent syntax. Constituency trees have the advantage of a well-defined linearization order compared to dependency trees. Further, constituency trees may be easier to realise, as they effectively correspond to a bracketing of the surface form.

Unfortunately, AMR annotated data does not come with syntactic annotation. We therefore parse the training and dev splits of both corpora with the Stanford parser \citep{Manning:14} to provide silver-standard reference parse trees. We then delexicalise the parse trees by trimming the trees of the surface words; after this stage, the leaves of the tree are the preterminal POS tags. After this, we linearise the delexicalised constituency trees with depth-first traversal, following \citet{Vinyals:15}.

\section{Model implementation and training}

\subsection{Model details}
We wish to estimate $P(Y, Z|X)$, the joint probability of a parse $Y$ and surface form $Z$ given an AMR graph $X$. We model this in two parts, using the chain rule to decompose the joint distribution. The first model, which we call the syntax model, approximates $P(Y|X)$, the probability of a particular syntactic structure for a meaning representation. The second is $P(Z|X, Y)$, the lexicalisation model. This calculates the probability of a surface realisation given a parse tree and an AMR graph. We implement both as recurrent sequence-to-sequence models.

As we are able to linearise both the AMR graph and the parse tree, we use LSTMs \citep{Hochreiter:97} both as the encoder and the decoder of our seq2seq models. Given an input sequence $X_1, \dots, X_n$, which can either be an AMR graph or a parse tree, we first embed the tokens to obtain a dense vector representation of each token $x_1, \dots, x_n$. Then we feed this into a stacked bidirectional LSTM encoder to obtain contextualised representations of each input token $c_i$. As far as possible, we share parameters between our two models. Concretely, this means that the syntax model uses the same AMR and parse embeddings, and AMR encoder, as the lexicalisation model. We find that this speeds up model inference, as we only have to encode the AMR sequence once for both models. Further, it regularises the joint model by reducing the number of parameters.

In our decoder, we use the dot-product formulation of attention \citep{Luong:15}: the attention potentials $a_i$ at timestep $t$ are given by \begin{equation*}
    a_i = h_{t-1}^T W_{att} c_i
\end{equation*}
where $h_{t-1}$ is the decoder hidden state at the previous timestep, and $c_i$ is the context representation at position $i$ given by the encoder. The attention weight $w_i$ is then given by a softmax over the attention potentials, and the overall context representation $s_t$ is given by $\sum w_i c_i$. The syntax model only attends over the input AMR graph; the linearisation model attends over both the input AMR and syntax tree independently, and the resulting context representation $s_t$ is given by the concatenation of the AMR context representation and the syntax tree context representation \citep{Libovicky:17}.

We use $s_t$ to augment the input to the LSTM: $\widetilde{y}_t = W_{in} \tanh( [y_t; s_t])$. Then the LSTM hidden and cell state are updated according to the LSTM equations: $h_t, c_t = LSTM(h_{t-1}, c_{t-1}, \widetilde{y}_t)$. Finally, we again concatenate $s_t$ to $h_t$ before calculating the logits over the distribution of tokens: 
\begin{align}
    \widetilde{h}_t &= \tanh(W_{out}[h_t; s_t]) \\
    p(y_t | y_{<t}) &= \softmax(W \widetilde{h}_t)
    \label{eqn:lex}
\end{align}
For the syntax model, we further constrain the decoder to only produce valid parse trees; as we build the parse tree left-to-right according to a depth-first traversal, the permissible actions at any stage are to open a new constituent, produce a terminal (i.e. a POS tag), or close the currently open constituent. We implement this constraint by setting the logits of all impermissible actions to negative infinity before taking the softmax. We find that this improves both training speed and final model performance, as we imbue the decoder with an intrinsic bias towards producing well-formed parse trees.

\subsection{Generation with a copy mechanism}

Despite the preprocessing procedures referred to in Section \ref{sec:data}, we found that the lexicalisation model still had trouble with out-of-vocabulary words, due to the small size of the training corpus. This led to poor vocabulary coverage on the unseen test portions of the dataset. On closer inspection, many out-of-vocabulary words in the dev split are open-class nouns and verbs, which correspond to concept nodes in the AMR graph. We therefore incorporate a copy mechanism \citep{Vinyals:15b,See:17} into our lexicalisation model to make use of these alignments.

We implement this by decomposing the word generation probability into a weighted sum of two terms. One is the vocabulary generation term. This models the probability of generating the next token from the model vocabulary, and is calculated in the same way as the base model. The other is a copy term, which calculates the probability of generating the next token by copying a token from the input. This uses the attention distribution over the input tokens calculated in the decoder to decide which input token to copy. The weighting between these two terms is calculated as a function of the current decoder input token, the decoder hidden state, and the AMR and parse context vectors. To sum up, the per-word generation probability in the decoder is given by
\begin{equation}
    p(y_t | y_{<t}) = (1 - \theta_t)p_{lex}(y_t | y_{<t}) + \theta_t \sum_{i : X_i = y_t} w_i
\end{equation}
where $p_{lex}(y_t | y_{<t})$ is as in Equation \ref{eqn:lex} and $w_i$ is the attention weight on the input token $X_i$. $\theta$ is the weighting between the generation term and the copy term: this is implemented as a 2-layer MLP.

\subsection{Model training procedures}

The AMR training corpus, together with the automatically derived parse trees, give us aligned triples of AMR graph, parse tree and realisation. We train our model to minimise the sum of the parse negative log-likelihood from the syntax model and the text negative log-likelihood from the lexicalisation model. We use the ADAM optimizer \citep{Kingma:15} with batch size 40 for 200 epochs. We evaluate model BLEU score on the dev set during training, and whenever this did not increase after 5 epochs, we multiplied the learning rate by 0.8. We select the model with the highest dev BLEU score during training as our final model.

We apply layer normalization \citep{Ba:16} to all matrix multiplications inside our network, including in the LSTM cell, and drop out all non-recurrent connections with probability 0.5 \citep{Srivastava:14}. We also drop out recurrent connections in both encoder and decoder LSTMs with probability 0.3, tying the mask across timesteps as suggested by \citet{Gal:16}. All model hidden states are size 500, and token embeddings are size 300. Word embeddings are initialised with pretrained \texttt{word2vec} embeddings \citep{Mikolov:13}. We replace words with count 1 in the training corpus with the UNK token with probability 0.5, and replace POS tags in the parse tree and AMR concepts with the UNK token with probability 0.1 regardless of count.

\paragraph{Decoding from our model}
During test time, we would like to estimate 
\begin{equation}
    \argmax_{Z} \sum_Y P(Z, Y | X)
    \label{sec:decoding}
\end{equation}
the most likely text realisation of an AMR, marginalising out over the possible parses. To do this, we heuristically find the $n$ best parses $Y_1, \dots, Y_n$ from the syntax model, generate a realisation $Z_i$ for each parse $Y_i$, and take the highest scoring parse-realisation pair as the model output.

We use beam search with width 2 for both steps, removing complete hypotheses from the active beam and appending them to a $k$-best list. We terminate search after a predetermined number of steps, or if there are no active beam items left. After termination, if $k > n$, we return the top $n$ items of the $k$-best list; otherwise we return additional items from the beam. In our experiments, we find that considering realisations of the 2 best parses (i.e. setting $n = 2$ above) gives the highest BLEU score on the dev set.

\section{Experiment 1: AMR and syntax}
\label{sec:parsing}
\begin{table}[t]
    \centering
    \begin{tabular}{ccc}
        \toprule
        Model &  Unlabelled F1 & Labelled F1\\
        \midrule
        Text-to-parse & 87.5 & 85.8 \\
        AMR-to-parse & 60.4 & 54.8 \\
        Unconditional & 38.5 & 31.7 \\
        \bottomrule
    \end{tabular}
    \caption{Parsing scores on LDC2017T10 dev set.}
    \label{tab:parsing}
\end{table}

\begin{table}[t]
    \centering
    \begin{tabular}{cc}
        \toprule
        Model & \# good realisations \\
        \midrule
        Syntax-aware model & 1.52 \\
        Baseline s2s & 1.19 \\
        \bottomrule
    \end{tabular}
    \caption{Average number of acceptable realisations out of 3. The difference is significant with $p< 0.001$.}
    \label{tab:paraphrase}
    \vspace{-1em}
\end{table}

\begin{table}[t]
    \centering
    \resizebox{\columnwidth}{!}{
    \begin{tabular}{lcc}
        \toprule
        Model & Dev BLEU & Test BLEU\\
        \midrule
        \multicolumn{3}{l}{\textit{Trained on LDC2017T10}} \\
        Our model & \textbf{26.1} & \textbf{26.8} \\
        Our model + oracle parse & 57.5 & - \\
        Baseline s2s + copy & 23.7 & 23.5 \\
        \citet{Beck:18} & - & 23.3 \\
        \multicolumn{3}{l}{\textit{Trained on LDC2015E86}} \\
        Our model & \textbf{23.6} & \textbf{23.5} \\
        Our model + oracle parse & 53.1 & - \\
        \citet{Konstas:17} & 21.7 & 22.0 \\
        \citet{Song:18} & 22.8 & 23.3 \\
        \multicolumn{3}{l}{\textit{Trained on LDC2015E86 or earlier + additional unlabelled data}} \\
        \citet{Song:18} & - & 33.0 \\
        \citet{Konstas:17} & 33.1 & 33.8 \\
        \citet{Pourdamghani:16} & 27.2 & 26.9 \\
        \citet{Song:17} & 25.2 & 25.6 \\
        \bottomrule
    \end{tabular}
    }
    \caption{BLEU results for generation.}
    \label{tab:bleu}
    \vspace{-1em}
\end{table}

We first investigate how much information AMR contains about possible syntactic realisations. We train two seq2seq models of the above architecture to predict the delexicalised constituency tree of an example given either the AMR graph or the text. We then evaluate both models on labelled and unlabelled F1 score on the dev split of the corpus. As neither model is guaranteed to produce trees with the right number of terminals, we first run an insert/delete aligner between the predicted and reference terminals (i.e. POS tags) before calculating span F1s. We also report the results of running our aligner on the most probable parse tree as estimated by an unconditional LSTM as a baseline both to control for our aligner and also to see how much extra signal is in the AMR graph. The results in Table \ref{tab:parsing} show that predicting a syntactic structure from an AMR graph is a much harder task than predicting from the text, but there is information in the AMR graph to improve over a blind baseline.

\section{Experiment 2: Generating natural language from AMR}
Table \ref{tab:bleu} shows the results of our model on the AMR generation task. We evaluate using BLEU score \citep{Papineni:02} against the reference realisations. As a baseline, we train a straight AMR-to-text model with the same architecture as above to control for the extra regularisation in our model compared to previous work. Our results show that adding syntax into the model dramatically boosts performance, resulting in state-of-the-art single model performance on both datasets without using external training data.

As an oracle experiment, we also generate from the realisation model conditioned on the ground truth parse. The outstanding result here -- BLEU scores in the 50s -- demonstrates that being able to predict the gold reference parse tree is a bottleneck in the performance of our model. However, given the inherent difficulty of predicting a single syntax realisation (cf. Section \ref{sec:parsing}), we suspect that there is an intrinsic limit to how well generating from an AMR graph can replicate the reference realisation.

We further note that we do not use models tailored to graph-structured data or character-level features as in \citet{Song:18,Beck:18}, or additional unlabelled data to perform semi-supervised learning \citep{Konstas:17}. We believe that we can improve our results even further if we use these techniques.

\section{Experiment 3: Generating varied realisations}
Our model explicitly disentangles variation caused by syntax choice from that caused by lexical choice. This means that we can generate diverse realisations of the same AMR graph by sampling from the syntax model and deterministically decoding from the realisation model. We hypothesise that this procedure generates more meaning-preserving realisations than just sampling from a straight AMR-to-text model, which can result in incoherent output \citep{Cao:17}.

We selected the first 50 AMR graphs in the dev set on linearised length between 15 and 40 with coherent reference realisations and generated 5 different realisations with our joint model and our baseline model. For our joint model, we first sampled 3 parse structures from the syntax model with temperature 0.3. This means we divide the per-timestep logits of the syntax decoder by 0.3; this serves to sharpen the outputs of the syntax model and constrains the sampling process to produce relatively high-probability syntactic structures for the given AMR. Then, we realised each parse deterministically with the lexicalisation model. For the baseline model, we sample 3 realisations from the decoder with the same temperature. This gave us 100 examples in total.

We then crowdsourced acceptability judgments for each example from 100 annotators: we showed the reference realisation of an AMR graph, together with model realisations, and asked each annotator to mark all the grammatical realisations which have the same meaning as the reference realisation. Each annotator was presented 30 examples selected randomly. Our results in Table \ref{tab:paraphrase} show that the joint model can generate more meaning-preserving realisations compared to a syntax-agnostic baseline. This shows the utility of separating out syntactic and lexical variation: we model explicitly meaning-preserving invariances, and can therefore generate better paraphrases.

\section{Conclusions and further work}

We present an AMR generation model that factors the generation process through a syntactic decision, and show that this leads to improved AMR generation performance. In addition, we show that separating the syntactic decisions from the lexicalisation decisions allows the model to generate higher quality paraphrases of a given AMR graph.

In future work, we would like to integrate a semantic parser into our model \citep{Yin:18}. Annotating data with AMR is expensive, and existing AMR treebanks are small. By integrating a component which parses into AMR into our model, we can do semi-supervised learning on plentiful unannotated natural language sentences, and improve our AMR generation performance even further. In addition, we would be able to generate text-to-text paraphrases by parsing into AMR first and then carrying out the paraphrase generation procedure described in this paper \citep{Iyyer:18}. This opens up scope for data augmentation for downstream NLP tasks, such as machine translation.

\section*{Acknowledgements}

The authors would like to thank Amandla Mabona and Guy Emerson for fruitful discussions. KC is funded by an EPSRC studentship.
    
\bibliography{naaclhlt2019.bib}
\bibliographystyle{acl_natbib}

\end{document}